# An efficient recommendation model based on Knowledge Graph Attention-assisted Network (KGAT-AX)

Zhizhong Wu [1, a *]

[1] UC Berkeley, Berkeley, US

[a] ecthelion.w@gmail.com

**Abstract.** Recommendation systems play a crucial role in helping users filter through vast amounts of information. However, traditional recommendation algorithms often overlook the integration and utilization of multi-source information, limiting system performance. Therefore, this study proposes a novel recommendation model, Knowledge Graph Attention-assisted Network (KGAT-AX). We first incorporate the knowledge graph into the recommendation model, introducing an attention mechanism to explore higher order connectivity more explicitly. By using multilayer interactive information propagation, the model aggregates information to enhance its generalization ability. Furthermore, we integrate auxiliary information into entities through holographic embeddings, aggregating the information of adjacent entities for each entity by learning their inferential relationships. This allows for better utilization of auxiliary information associated with entities. We conducted experiments on real datasets to demonstrate the rationality and effectiveness of the KGAT-AX model. Through experimental analysis, we observed the effectiveness and potential of KGAT-AX compared to other baseline models on public datasets. KGAT-AX demonstrates better knowledge information capture and relationship learning capabilities.

## 1. Introduction

Recommendation systems are technologies used to provide personalized recommendations to users. However, traditional recommendation systems face challenges such as data sparsity, cold start, and difficulties with personalized recommendations. In recent years, knowledge graph-based recommendation systems have emerged as a promising solution to these issues and have gained widespread attention.

A knowledge graph is a method of representing knowledge using a graph structure. It connects entities, attributes, and relationships in a structured manner, constructing a rich network of knowledge. This network not only contains vast amounts of data, but also exhibits semantic relationships between entities. This knowledge can be utilized to provide more accurate and personalized recommendations, thereby enhancing the performance and user experience of recommendation systems.

Knowledge graph-based recommendation systems leverage this rich network of knowledge to offer more accurate and personalized recommendations. They can gain a deeper understanding of user needs and preferences and consider relationships between different entities based on user interests, behaviors, and contextual information. This enables the system to present recommendations with greater depth and breadth. Knowledge graph-based recommendation systems have been widely applied in various domains, including e-commerce, social media platforms, video and music streaming, and more. In summary, knowledge graph-based recommendation systems overcome the limitations of traditional recommendation systems and provide precise and personalized recommendations. As knowledge graphs continue to evolve, they will play an increasingly important role in various domains, offering users with an enhanced recommendation experience.

Currently, traditional knowledge graph-based recommendation methods include embedding-based approaches and path-based approaches. Embedding-based methods represent semantic associations by learning low-dimensional embedding vectors for entities and relationships. Common models include TransE, TransH, and others [1]. Path-based methods infer relational connections by analyzing path information between entities. Common models include Personalized PageRank and PathSim [2]. However, these methods face challenges in dealing with large-scale graphs, dynamic graphs, and computational efficiency. Therefore, researchers often combine the advantages of these methods to improve recommendation effectiveness.

While the research field of knowledge graph-based recommendation systems is becoming increasingly active and attracting extensive exploration and attention, there are still challenges in knowledge representation learning, dynamic knowledge graph modeling, cold start, and interpretability. Thus, this paper proposes an innovative approach, the Knowledge Graph Attention Network with Auxiliary Information, for recommendation systems. This approach aims to enhance the effectiveness of knowledge graph-based recommendations by leveraging auxiliary information. It can adapt flexibly to changes in dynamic knowledge graphs through the completion of auxiliary information and providing reliable recommendations in cold start scenarios.

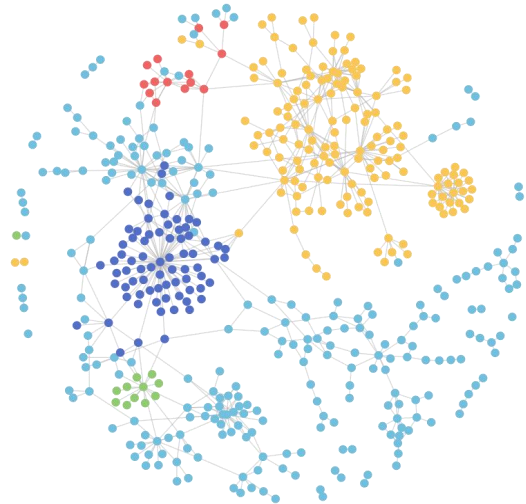

Figure 1. Knowledge Graph demonstration diagram

## 2. BASIC THEORY

### 2.1 Knowledge graph

Knowledge graph is a structured representation method used to describe relationships between entities, such as people, places, events, etc. It is a graph data model consisting of nodes and edges, where nodes represent entities and edges represent relationships between entities. The goal of the knowledge graph is to organize a large amount of information and knowledge so that machines can understand and reason with it. By connecting related entities and properties, it forms a semantic network that provides rich relational and contextual information. [11]

The construction of a knowledge graph typically involves extracting and integrating information from both structured and unstructured data sources. This data can come from various sources such as text documents, databases, the Internet, etc.

Through techniques like natural language processing, information extraction, and entity linking, knowledge graphs can automatically extract and organize this information, forming a knowledge network with semantic connections.

Knowledge graphs have widespread applications in various fields, including semantic search, intelligent question-answering, recommendation systems, natural language processing, etc. They provide machines with a deeper and more comprehensive understanding capability, helping machines understand and reason about relationships between entities, thus supporting more intelligent applications and services.

In summary, a knowledge graph is a graphical structure used to represent and organize knowledge. It provides rich descriptions of relationships between entities, serves as a foundation for machine understanding and reasoning, and promotes the development of artificial intelligence.

Graph Attention Neural Network [5] is a neural network model used to process graph data. Unlike traditional neural networks, graph attention neural networks specialize in handling graph-structured data with nodes and edges, which isvery common in many practical problems such as social networks, recommendation systems, chemical molecules, etc.

**2.2 Graph Attention Neural Network**

The core idea of Graph Attention Neural Network is to weigh different nodes in a graph using an attention mechanism. It can learn the relationship strength between each node and its neighboring nodes and dynamically allocate attention weights based on the relevance between nodes. [14] This allows the Graph Attention Neural Network to adaptively focus on nodes with greater influence or importance.

In Graph Attention Neural Network, each node has a feature vector, and edges represent connections between nodes. Through multiple layers of graph convolution operations, the network can perform information propagation and aggregation on node features. [12] The graph attention mechanism enables the network to dynamically weight the information propagation process based on the relationships between nodes. [13]

Graph Attention Neural Network has been widely applied in tasks such as graph classification, node classification, and graph generation, achieving excellent performance in these domains. It effectively captures both local structures and global relationships in a graph, providing a powerful tool for processing complex graph-structured data.

In summary, the Graph Attention Neural Network is a neural network model specifically designed to process graph data. It uses an attention mechanism to weigh the relationships between nodes, making it valuable in various graph-related tasks.

## 3. Method

We propose a KGAT-AX model that uses the entire high-level relationship. The model framework is shown in Figure 2 and consists of three main components: 1) Embedding layer: This layer parameterizes each node as a vector while preserving the structure of the CKG (Complementary Knowledge Graph); 2) Embedding propagation layer: This layer recursively propagates the embeddings of nodes' neighbors to update their representations. It uses a data-based concern mechanism to calculate the e-weights of each neighbor during the propagation process. 3) Prediction layer: This layer combines user and element representations from all propagation layers and generates predicted match values.

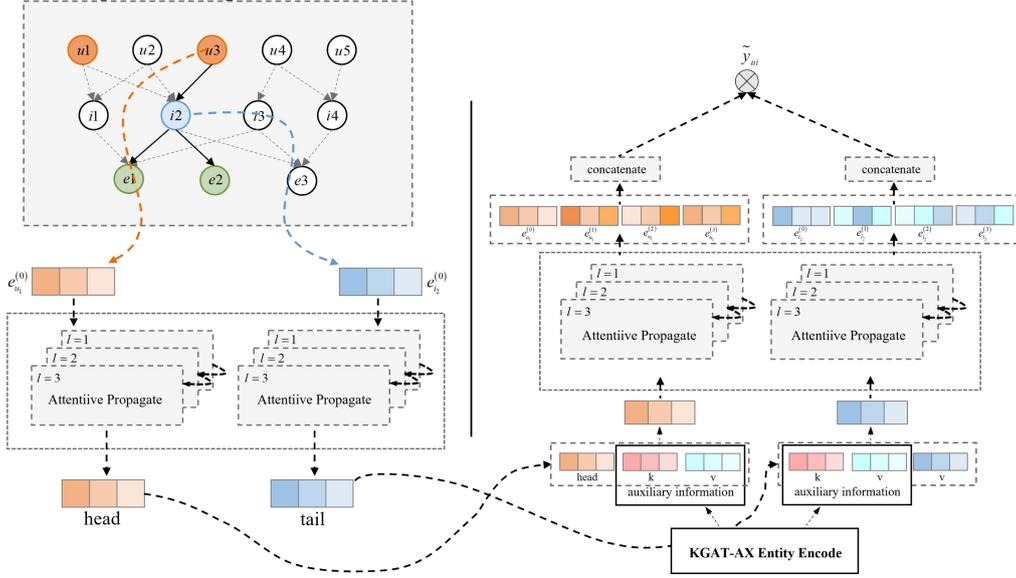

Figure 2. Schematic architectural diagram of the KGAT-AX entity encoding and propagation process

### 3.1 Embedding Layer

The task of the data graph embedding layer is to transform the symbolic entities and relationships of the data graph into continuous vector representations, which allow semantic information to be expressed numerically. This increases the meaningfulness of the data graph and its application potential.

The embedding layer model that is often used in recommendation systems is TransR [6]. Specifically, it maps entities and relationships into distinct vector spaces represented by two embedding vectors: one for the entity space and another for the relationship space. For each triangle (h, r, t), the TransR model changes it to (uh, er, et). The following basic principles are:

Therefore, for a given triangle (h, r, t), the credibility score is given as:

$$g(h,r,t) = \|w_r e_h + e_r - w_r e_t\|_2^2 \qquad (1)$$

Below, Wr represents the transformation matrix for relation r that describes the entities from the origin space to the entity space and relation space. The score g(h,r,t) measures the correspondence between the main entity h, relation r, and tail entity t. In TransR, a smaller value of g(h,r,t) indicates a higher probability that the triple is true.

During the TransR training process, the relative order between valid and defective triangles is taken into account, and pairwise ranking loss is used to drive their differentiation. The aim is to learn embeddings that can better distinguish valid triangles from defective triangles. In particular, for certain triangles (h, r, t ) consider valid triples (h', r, t') and defective triples (h', r, t'), where h' and t' are randomly selected entities, with the aim of minimizing valid triple scores while maximizing the broken triple scores, with margin between them. This drives the model to assign lower scores to valid triangles and higher scores to corrupt triangles. The goal of the training process is to optimize the embedding and transformation matrices to achieve better discrimination between valid and corrupt triangles.

$$\zeta KG = \sum_{(h,r,t,t') \in \zeta} -\ln \sigma(g(h,r,t')) - g(h,r,t)) \qquad (2)$$

The proposed model considers a set of T-triplets containing valid triplets and corrupt triplets generated randomly using surrogate entities. The triplets are modeled using the sigmoid function o(.) and the connectivity is coupled directly into the representation. This layer acts as a regulator, which improves the performance of model representation [4].

Here T = {(h, r, t, t') | (h, r, t)∈G, (h, r, t')∉ G}, where G represents the set of valid triplets and (h, r, t') is a broken triplet constructed by randomly replacing one entity from a valid triplet. The sigmoid function o(.) is used to model this triple, and the contact information is injected directly into the representation to increase the power of the model representation.

**3.2 Propagation layer**

When studying the embedding graph for candidate entity h, two aspects must be taken into account. First, the structured representation of the data graph is studied using the transE model, where the representation is denoted by h+st. Second, to enhance the representation of entity h, information on multimodal neighboring entities is combined into h. Following the approach proposed in [1], Nh represents the set of triplets directly connected to h. The vector representation of eagg obtained by linearly combining each triplet representation can be calculated using the equation 1.

$$e_{agg} = \sum_{(h,r,t)\in N_h} \pi(h,r,t) e(h,r,t) \qquad (3)$$

The embedding representation of each triplet (h, r, t) is denoted by e(h, r, t) and the point of interest of each triplet e(h, r, t) is denoted by Tt(h, r, t). The point of interest T(h, r, t) ) determines how far the information spreads out from the triplet e(h, r, t).

To understand the importance of relationships in graphing data, we introduce learnable parameters to embed the relationships e(h, r, t) and n(h, r, t). For the triplet e(h, r, t), we learn the insertion by performing a linear transformation with the concatenating of the head entity, tail entity, and insertion of the relation as shown below:

$$e(h,r,t) = w_1 (e_h \| e_r \| e_t) \qquad (4)$$

The embedding of the head entity and embedding of the tail entity are represented by eh and et respectively, while the embedding of the relationship is represented by er. To calculate n(h, r, t) using the relational attention mechanism, it can be expressed as:

$$\pi(h,r,t) = Leaky \operatorname{Re} LU(w_2 e(h,r,t)) \qquad (5)$$

To maintain consistency with previous work [10], we adopted the activation function LeakyReLU [15] as the non-linear activation function. We then used the softmax function to normalize the coefficients of all the triangles connected to the entity h.

$$\pi(h,r,t) = \frac{\exp(\pi(h,r,t))}{\sum_{(h,r',t')\in N_h} \exp(\pi(h,r,t'))} \qquad (6)$$

**3.3 Prediction layer**

After the data graph embedding module, each entity gets the appropriate embedding, which is then fed into the recommendation module. Similar to the data graph embedding module, to store 1-hop to n-hop information, we use the method proposed in [3]. This method stores the output of potential users and targets in the 1st layer. Each output from a layer represents information from a different jump. To combine the representations of each step into a single vector, we use a layer aggregation mechanism [7]. The aggregation process can be described as follows:

$$e_u^* = e_u^{(0)} \| \cdots \| e_u^{(L)}, e_i^{(*)} = e_i^{(0)} \| \cdots \| e_i^{(L)} \quad (7)$$

The representation of each MKG concern layer is combined using a cascade of 7 operations, denoted L, where L represents the number of layers. This approach, as described in [15], enriches the original embedding by embedding the propagation function and allows the propagation strength to be adjusted by adjusting L.

Finally, we calculate the similarity score between the user and product representations by running the product in as shown in Equation 8 to predict its compatibility.

$$\hat{y}(u,i) = e_u^{*T} e_i^* \quad (8)$$

Furthermore, we optimized the prediction loss of the recommendation model using Bayesian Personalized Ranking (BPR) loss [8]. This loss gives a higher prediction score for observed records, indicating a stronger user preference compared to undetected records. BPR losses can be determined mathematically by Equation 9.

$$\zeta_{recsys} = \sum_{(u,i,j) \in o} -\ln \sigma \left( \hat{y}(u,i) - \hat{y}(u,j) \right) + \lambda \|\Theta\|_2^2 \quad (9)$$

In the equation, O = {(u, i, j) | (u, i) ∈ R+ and (u, j) ∈ R-} represents the training set, where R+ stands for the observed interaction between user u and object i and R- represent unobserved sample interactions. σ (.) represents sigmoid function. O represents parameter set and θ represents parameter with L2 regularization [9][10].

We repeatedly update the parameters of the embed module and the MKG recommendations module. Specifically, for a randomly selected pool (h, r, t, t'), we update the embedding graph of data from all entities. Then we randomly sample the pool ( u, i, j) and retrieve their representation from the embedded data graph. The loss functions of the two modules are optimized in turn.

### 3.4 Fusion layer

The holographic embeddings of entities and relations are established using mathematical mappings into a lower-dimensional space. Simultaneously, auxiliary information, encompassing contextual attributes, is integrated into the framework. This process involves element-wise multiplication achieved through the Hadamard product between the holographic embedding of a subject entity and the auxiliary information corresponding to the subject and predicate. This operation effectively merges the semantic information encoded in both the original triplet and the auxiliary data, generating augmented triplets. The augmented triplets manifest in the following forms: (Head Entity, Auxiliary Information, Auxiliary Information), (Relation, Auxiliary Information, Auxiliary Information), and (Tail Entity, Auxiliary Information, Auxiliary Information).

Subsequently, we will iteratively revisit the propagation layer's procedure.

## 4. Experiment

In this section, we detail our experimental process and report experimental results to verify model performance. In this section, we evaluate the performance of the KGAT-AX model on two real data sets (movies and music).

## 4.1 Datasets

The public dataset we use is MovieLense20.

The public datasets we use are MovieLens. Detailed data are shown in Table 1.

Table 1. Data set related data presentation

| classification | MovieLens20 |
|---|---|
| User ID | 138493 |
| Movie ID | 27278 |
| Film title | 27278 |
| Types of films and their labels | 67 |
| Mark | 20000263 |
| Film official label | 11709769 |
| relevance | 465564 |
| The user's tag for the movie | 1128 |
| Movie tag ID | |

For each data set, we randomly select 80% of each user interaction history to form the training set, and treat the rest as a test set. From the training set, we randomly select 10% of interactions as a validation set to set the hyperparameters. We treat each item-user interaction detected as a positive example, then employs a negative sampling strategy to associate it with a negative item that has not been used by the user before.

## 4.2 Parameter Settings

We use Xavier to initialize users, items, attributes, parameters, etc. and optimize all models with Adam opti mizer. In order to reduce the time complexity, firstly, the grid search method is used to determine the optimal value of each super parameter in the model. The search range is set as fol lows: batch size = {128, 256, 512, 1024}, dimension D = {8, 16, 32, 64}, learning rate range lr={1e−6, 1e−5, 1e− 4, 1e−3}, number of sampled neighbors N={10, 20, 25, 50}, and depth H={1, 2, 3}. In addition, in order to improve the generalization ability of the model and alleviate overfitting, dropout, L2 regularization and earlyStopping strategies are added during training.

We implement our KGAT model in Tensorflow. The embedding size is fixed to 64 for all models, except RippleNet 16 due to its high computational cost. We optimize all models with Adam optimizer, where the batch size is fixed at 1024.The default Xavier initializer to initialize the model parameters.We apply a grid search for hyperparameters: tuned learning rates among (0.05, 0.01, 0.005, 0.001), searched L2normalization coefficient in (10-5,10-4,..., 101, 10), and tuned dropout ratio in (0.0, 0.1,..., 0.8) for NFM, GC-MC, and KGAT.

## 4.3 Experimental analysis

4.3.1 Performance comparison: Performance comparison results are presented in Table I. We have the following observations:Table styles

Table 2. Overall performance comparison

| Model | Movielens20 | | Last-FM | |
|---|---|---|---|---|
| | recall | ndcg | recall | ndcg |
| FM | 0.1139 | 0.2538 | 0.1034 | 0.1172 |
| MF | 0.1194 | 0.4342 | 0.1036 | 0.1402 |
| NeuralKG[18] | 0.1192 | 0.4283 | 0.1037 | 0.1367 |
| CFKG | 0.1214 | 0.2635 | 0.0962 | 0.1099 |
| KGCN | 0.1127 | 0.4286 | 0.1074 | 0.1406 |
| KGAT-AX | 0.1296 | 0.4371 | 0.1095 | 0.1467 |
| %Improve | +5.87% | +2.02% | +1.81% | +4.36% |

The performance comparison results are shown in Table III, and we can draw the following conclusions: KGAT-AX achieves the best performance compared to other baseline models, with an increase of x,x in recall2.02% on the public dataset. By incorporating auxiliary information and forming new triplets through holistic embeddings, KGAT-AX is able to capture knowledge information more effectively and enhance its relational learning capabilities.

CFKG [19] and KGCN [20] perform poorly on the Movielens20 dataset. This is because CFKG relies on regularization methods that require strict connectivity, limiting its ability to make full use of item knowledge. As a result, MF [17] and FM [16] outperformed CFKG and KGCN in performance. In contrast, KGAT-AX, with multiple stacked attention propagation layers, can explore high-order connectivity in a more explicit way, effectively capturing collaborative signals. The overall superior performance of KGAT-AX compared to other baseline models is mainly due to the holistic embeddings of auxiliary information, which form two new triplets with the original triplet heads and tails. This provides richer semantic and contextual information, improves the quality and accuracy of knowledge representation, and makes more effective use of information in the knowledge graph. It enriches the semantics of entities and relationships and makes full use of the information in the graph.

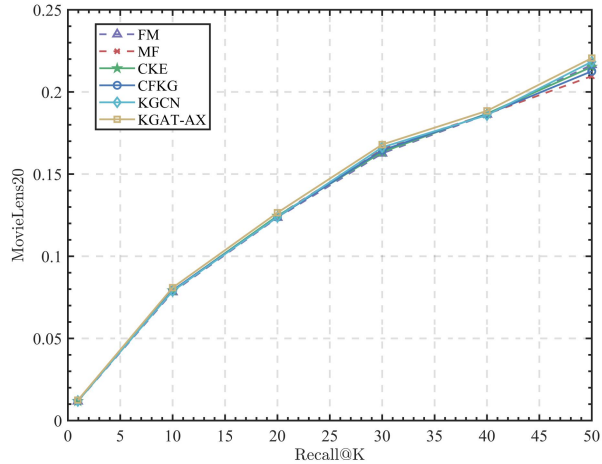

Figure 3. Comparison of Recall@K performance of different recommendation algorithms on the MovieLens20 data set

In order to verify the influence of the KGAT-AX model in the KG embedding and interactive attention mechanism, we still use KGAT-AXno ConvE , KGAT-AXnoAtt eu , KGAT-AXnoAtt ei and KGAT-AX for comparative experiments to verify the effec tiveness of KGAT-AX.As shown in Table 4:

Table 3 Auc performance kgat-ax variants ablation experiments

| Model | MovieLense20 |
|---|---|
| KGAT-AXno_ConvE | 0.9821 |
| KGAT-AXno_Att_eu | 0.9819 |
| KGAT-AXno_Attei | 0.9791 |
| KGAT-AX | 0.9831 |

Our experimental variants, including KGAT-AXno ConvE, KGAT-AXnoAtt eu, KGAT-AXnoAtt ei, and KGAT-AX, were trained using the same hyperparameter settings as KGAT-AX. Analyzing the experimental results presented in Table X, we draw the following conclusions: 1) Removing the embedding of the pre-trained knowledge graph leads to a decrease in the performance of the KGAT-AX model. This reduction can be attributed to the fact that KG embedding provides crucial first-order structural information during the pre-training phase. 2) The removal of the interactive attention network between the user-side representation and the item-side representation significantly reduces KGAT-AX performance. This finding highlights the importance of treating multi-order neighbor information equally, as it strongly influences the model's performance. Incorporating neighbor weight information when modeling the user interactively with the project results in improved recommendations.

## 5. Summary

This paper proposes a new recommendation model based on a knowledge graph called Knowledge Graph Attention eXplicit (KGAT-AX), which integrates incremental information into entity embedding using holistic embedding. The KGAT-AX model exploits inference relationships between entities and collects information from neighboring entities to within each entity, enabling each entity to make better use of the additional information.Extensive experiments on two real-world datasets demonstrated the rationality and effectiveness of the proposed KGAT-AX model.This work is a study on integrating additional information into a holistic vector using a holistic embedding approach in recommendation system, which enables better integration and utilization of multi-source information. It also provides an opportunity for future researchers to study the holistic application possibilities of the embedding method in recommendation systems with the help of additional tests and validations.